\documentclass[runningheads]{llncs}

\usepackage{eccv}

\usepackage{eccvabbrv}
\usepackage{graphicx}
\usepackage{booktabs}
\usepackage{amsmath}
\usepackage{amssymb}
\usepackage{multirow}
\usepackage[accsupp]{axessibility}  
\usepackage{url}
\usepackage{hyperref}
\usepackage{orcidlink}
\makeatletter
\g@addto@macro\normalsize{%
  \setlength{\abovedisplayskip}{5pt plus 2pt minus 2pt}%
  \setlength{\belowdisplayskip}{5pt plus 2pt minus 2pt}%
  \setlength{\abovedisplayshortskip}{1pt plus 2pt}%
  \setlength{\belowdisplayshortskip}{4pt plus 2pt minus 2pt}%
}
\makeatother

\begin{document}

\title{RootQuantV2: Adapting a Vision Foundation Model for Root-Trait Regression from Minirhizotron Imagery}
\titlerunning{RootQuantV2: Adapting a Foundation-Model Root-Trait Regression}
\author{Kinjalk Parth\inst{1} \and
Sebastian Varela\inst{2} \and
Andrew D.B. Leakey\inst{1,2}}
\authorrunning{K.~Parth et al.}
\institute{Department of Plant Biology, University of Illinois at Urbana-Champaign, Urbana, IL 61801, USA\\
\email{kinjalk2@illinois.edu, leakey@illinois.edu} \and
Center for Advanced Bioenergy and Bioproducts Innovation, Urbana, IL 61801, USA\\
\email{sv79@illinois.edu}}
\maketitle

\begin{abstract}
A lack of high-throughput phenotyping solutions for root traits in field-grown crops has severely constrained understanding and improvement of below-ground traits and processes. Minirhizotrons are the standard non-destructive root-phenotyping method in field environments. Computer vision solutions are needed to allow automated trait estimation at scale, but training data is scarce and human annotations are often inaccessible because they reside in proprietary software that only exports per-image scalar totals of root \emph{length} and \emph{surface area}. Nevertheless, large numeric archives of these root traits already exist. RootQuant showed that the traits can be predicted directly from the whole image by regression, thus removing manually traced masks from the pipeline; RootQuantV2 takes that idea further by replacing RootQuant's CNN backbone with a self-supervised ViT. We adapt a frozen DINOv3 ViT-L/16 with a hybrid parameter-efficient scheme. Training only $11.9$M parameters ($3.78\%$ of the model), RootQuantV2 achieves length and area $R^2$ of $0.950$ and $0.930$, respectively, while lowering length/area RMSE by $24.3\%/20.7\%$ over RootQuant. RootQuantV2 thus repurposes legacy numeric archives for high-throughput, automated root trait estimation.

\keywords{Minirhizotron root phenotyping \and Segmentation-free trait regression \and
Vision foundation models \and Parameter-efficient fine-tuning \and
Self-supervised ViT adaptation}
\end{abstract}

\section{Introduction}
\label{sec:intro}

Root system traits are key crop improvement targets because they influence water and nutrient acquisition~\cite{lynch2013,lynch2019,bengough2011,wasson2014}. Minirhizotrons---transparent tubes installed in soil to depths of $1$\,m or more, and imaged repeatedly in the field---provide the principal non-destructive method for monitoring root dynamics over time and generate extensive image archives. Most commonly, expert annotators manually trace each image using proprietary software~\cite{winrhizo}, which stores the complete annotation, including pixel masks and skeletons, in an encoded internal format and exports only per-image scalar measurements such as living root length and surface area. These exported scalars constitute the gold standard for minirhizotron root quantification; consequently, a model trained on them reproduces the measurements needed for biological analysis. However, classical thresholding-and-skeletonization pipelines~\cite{lobet2017,seethepalli2021} generate substantial false positives across heterogeneous soils and imaging systems~\cite{baykalov2023}, whereas supervised segmentation methods require pixel-level annotations that are not available from proprietary exports~\cite{bauer2022,smith2022_rootpainter,wang2019_segroot}.

RootQuant~\cite{rootquantv1} addressed this limitation by reframing the problem as direct scalar-trait regression. Rather than predicting a spatially resolved segmentation mask, RootQuant employs an ImageNet-pretrained InceptionResNet-V2 backbone~\cite{szegedy2017} with a lightweight regression head to estimate root traits directly from the full image. This formulation enables learning from numeric-only archives and provides superior generalization ability across two major crop species compared to a single-species model. Because minirhizotron images contain fine, elongated root structures embedded within heterogeneous soil backgrounds~\cite{johnson2001,baykalov2023}, convolutional backbones may be limited by their predominantly local receptive fields, whereas ViTs can model long-range spatial dependencies through self-attention~\cite{dosovitskiy2021vit,simeoni2025dinov3}. Although RootQuant used a supervised convolutional backbone, recent advances in self-supervised vision transformers~\cite{caron2021dino,oquab2024dinov2,simeoni2025dinov3} may provide more transferable representations that improve generalization across species (\cref{sec:exp-xspecies}).

Three characteristics of the task and of modern self-supervised backbones guide our design. First, DINO-family self-supervised ViTs learn dense patch-level features that transfer to dense prediction tasks, including depth regression, without backbone fine-tuning~\cite{caron2021dino,oquab2024dinov2,simeoni2025dinov3}, so a frozen backbone with a lightweight task-specific head is attractive. Second, a standard ViT lacks an explicit two-dimensional spatial inductive bias, and purely linear low-rank adaptation does not introduce one; in contrast, convolutional adapters restore locality and have demonstrated strong performance on dense vision tasks~\cite{yin2024mona,jie2022convpass,yin2023lorand}. Third, \emph{length} and \emph{surface area} are extensive quantities that accumulate with image content rather than average over it. The natural design therefore sums local evidence additively and runs inference that preserves the exact symmetries of these quantities.

Guided by these considerations, we develop RootQuantV2 around three methodological contributions.
\begin{enumerate}
\item \textbf{Adaptation of a vision foundation model to root-trait regression.} We adapt a frozen DINOv3 ViT-L/16 backbone to whole-image regression of \textit{root length} and \textit{surface area}, replacing the convolutional backbone used in RootQuant. Segmentation-free CNN regression of root length has been implemented previously~\cite{rootquantv1,khoroshevsky2024_pp,khoroshevsky2024_cea}, and self-supervised backbones have been applied to plant-phenotyping tasks such as leaf counting~\cite{chen2023plantfm}; however, we are not aware of prior work adapting a self-supervised vision foundation model to minirhizotron root-trait regression. 
\item \textbf{A hybrid DoRA--Mona adaptation of a frozen ViT.} We combine DoRA~\cite{liu2024dora} on attention projections with a Mona multi-scale convolutional adapter~\cite{yin2024mona} on the MLP branch. The linear update mixes channels at attention, while the convolutional update restores locality during token-grid processing. Only $11.9$M parameters ($3.78\%$) are trained while the ViT backbone remains frozen. A matched ablation study demonstrates the contribution of the Mona branch, yielding a $9.0\%$ reduction in visible-root area RMSE without test-time augmentation and $5.5\%$ with it (\cref{sec:exp-ablation}).

\item \textbf{An extensive readout for image-wide accumulation of root evidence.} Length and surface area are extensive quantities that accumulate with image content rather than averaging over it, whereas conventional pooling operators are intensive. We therefore read the traits out by summing a per-patch density over the valid grid~\cite{lempitsky2010counting,zhang2016mcnn,liang2022transcrowd,zaheer2017deepsets}, and blend that sum with a pooled representation that concatenates the CLS token, a learnable attention pool, and GeM~\cite{radenovic2019gem}.
\end{enumerate}
On the visible-root subset---the $4{,}618$ test frames that contain roots---RootQuantV2 lifts combined $R^2$ (the mean of the length and area $R^2$) from $0.858$ to $0.914$ ($+6.5\%$ over RootQuant), and on the full test set to $0.940$ ($+4.4\%$, with both models evaluated on identical images). We report visible-root and full-set metrics side by side, and \cref{sec:exp-ablation} attributes most of the full-set gain to empty-frame false-positive suppression, with a smaller contribution from visible-root accuracy.
Code and trained weights are available at \url{https://github.com/leakey-lab/RootQuantV2}.

\vspace{-1em}
\section{Related Work}
\label{sec:related}

\subsubsection{Deep learning for root and minirhizotron imagery.}
Root phenotyping is dominated by dense prediction, segmenting the root and then measuring it~\cite{smith2020,smith2022_rootpainter,yasrab2019,wang2019_segroot,bauer2022}, which requires pixel- or skeleton-level annotations that are unavailable in numeric-only archives. Trait-estimation pipelines typically combine a U-Net or encoder--decoder segmenter with a trait calculator to recover root length, surface area, and diameter~\cite{wang2019_segroot,smith2020,bauer2022}, and therefore depend on proprietary tracing software or densely annotated labels for supervision~\cite{moller2019,atkinson2019,jiang2020,johnson2001}. Corrective-annotation tools reduce labeling effort~\cite{smith2022_rootpainter}, tip and skeleton extraction methods recover structural information~\cite{yasrab2019,pound2017}, benchmark datasets standardize evaluation~\cite{baykalov2023}, and domain-adaptive segmenters address temporal and cross-site distribution shifts~\cite{banet2024rootdomainshift}. The work most closely related to ours is direct trait regression from whole images~\cite{khoroshevsky2024_pp,khoroshevsky2024_cea}. RootQuant~\cite{rootquantv1} extended this approach to joint regression of root length and surface area using supervised convolutional networks. Unlike segmentation-based pipelines~\cite{wang2019_segroot,smith2022_rootpainter,bauer2022}, RootQuantV2 requires no pixel supervision, and unlike prior direct-regression approaches it adapts a self-supervised vision foundation model, replacing both the convolutional backbone and readout of RootQuant~\cite{rootquantv1}.

\subsubsection{Vision foundation models.}
Self-supervised ViTs produce dense patch-level representations that transfer effectively to downstream prediction tasks while the backbone remains frozen, and the DINO family~\cite{caron2021dino,oquab2024dinov2,simeoni2025dinov3} has progressively strengthened this property. DINO demonstrates that self-supervised attention can induce object segmentation without labels~\cite{caron2021dino}; DINOv2 provides general-purpose image- and pixel-level descriptors that remain effective when frozen~\cite{oquab2024dinov2}; DINOv3 further improves dense feature fidelity~\cite{simeoni2025dinov3}, and register tokens stabilize patch feature maps during inference~\cite{darcet2024registers}. Under a matched frozen-backbone protocol, DINOv2 features outperform OpenCLIP and MAE on monocular depth estimation~\cite{oquab2024dinov2}. Depth estimation is dense per-pixel prediction rather than the image-level scalar regression considered here, but these results demonstrate the transferability of self-supervised ViT representations. Evidence in agricultural imaging is more limited, but self-supervised pretraining has generally produced modest yet consistent improvements over ImageNet transfer, with the largest gains observed in low-label regimes~\cite{fomo4wheat2025,alnahian2025agrifm,sornapudi2024simclrag}. The closest precedent to our work is Chen~\etal~\cite{chen2023plantfm}, who adapt frozen foundation models with lightweight modules for plant-phenotyping tasks such as leaf counting.
\subsubsection{Parameter-efficient fine-tuning.}
On dense vision tasks, convolutional and multi-branch adapters consistently outperform purely linear low-rank tuning. Linear methods are the standard baseline: LoRA~\cite{hu2022lora} and its weight-decomposed variant DoRA~\cite{liu2024dora} adapt frozen weights with low-rank updates~\cite{han2024peftsurvey}, but they do not introduce explicit spatial structure. Convolutional adapters address this limitation by restoring locality. Mona, a multi-scale depthwise-convolution adapter, is the only delta-tuning method reported to surpass full fine-tuning on instance segmentation, semantic segmentation, and oriented object detection, while linear low-rank methods under the same protocol perform less strongly~\cite{yin2024mona}. ConvPass~\cite{jie2022convpass}, LoRand~\cite{yin2023lorand}, Conv-Adapter~\cite{chen2022convadapter}, and an input-conditioned convolutional adapter that nearly matches full fine-tuning on monocular depth regression~\cite{iconformer2024} show a similar ordering. Guided by this evidence, and by recent analyses of where adapters should be placed~\cite{steitz2024adapters}, we combine DoRA on the attention projections with a Mona adapter on the MLP branch.
\subsubsection{Pooling and inference for extensive quantities.}
Extensive targets, which grow with image content, are naturally read out by summing local evidence rather than averaging it. Density-based counting methods make this explicit by predicting a per-location density map and integrating it to obtain a total count~\cite{lempitsky2010counting,zhang2016mcnn,li2018csrnet}, while transformer-based counting models often regress the image-total directly~\cite{liang2022transcrowd}, and permutation-invariant set pooling formalizes summation as the aggregator for set-valued inputs~\cite{zaheer2017deepsets}. Generalized-mean (GeM) pooling, in contrast, interpolates between average and max pooling through a learnable exponent~\cite{radenovic2019gem}. Two training-free stabilization strategies complement such a readout. Test-time augmentation averages predictions over input transformations~\cite{shanmugam2021tta,lyzhov2020greedytta}; when the target is exactly invariant to those transformations, it can reduce prediction variance without introducing label inconsistency~\cite{wang2019tta,ashukha2020pitfalls,kimura2021tta}, although most reported gains have been observed in classification and segmentation rather than scalar regression.

\vspace{-1em}
\section{Methods}
\label{sec:method}

\subsection{Problem and data}
\label{sec:method-data}
We predict two scalars per RGB minirhizotron image: root length $\ell$ (mm) and surface area
$a$ (mm\textsuperscript{2}), regressed using only the numeric exports of proprietary tracing software~\cite{winrhizo}
and no pixel-level supervision, as described in RootQuant~\cite{rootquantv1}. Both targets are derived from a human expert's manual tracing and exported as scalar measurements from the software. The model therefore learns to reproduce the standard used for quantitative analysis, and all reported errors measure the agreement
with these expert annotations. We conduct all experiments using the
RootQuant dataset~\cite{rootquantv1}, which contains maize and soybean minirhizotron frames split into $89{,}186$ training,
$11{,}445$ validation, and $17{,}560$ test images. From the test split we remove $247$ frames with unreliable labels
($246$ labelled root-free that visual review confirmed to contain roots, and one frame whose area label is grossly
inconsistent with its image), leaving $17{,}313$ evaluated test images. The same removal is applied identically to
every model compared here. Frames are assigned to splits by stratified random sampling on the root-length distribution, with root-free frames as their own stratum, so the zero inflation is preserved in every partition~\cite{rootquantv1}. The labels are strongly zero-inflated. Among the $17{,}313$
test images, $12{,}692$ ($\sim\!73\%$) contain no root, $4{,}618$ contain roots, and the remaining $3$ images have only one nonzero trait. We define a per-image presence indicator
$m = \mathbf{1}[\ell>0 \wedge a>0]$, which is used for target standardization, the presence-balancing
loss weight (\cref{eq:loss}), and the visible-root evaluation subset (\cref{sec:exp-main}).
Because empty images dominate the dataset, standardizing over all images would bias the scale toward the empty majority and reduce the present-root signal. We therefore standardize each target $y\in\{\ell,a\}$ with the mean $\mu_y$ and standard
deviation $\sigma_y$ computed over the \emph{present} root samples ($m{=}1$) only,
\begin{equation}
z = \frac{y-\mu_y}{\sigma_y}, \qquad z_0 = -\frac{\mu_y}{\sigma_y}, \qquad
\hat{y} = \max\!\big(0,\ \hat{z}\,\sigma_y + \mu_y\big).
\label{eq:standardize}
\end{equation}
Each empty image ($y{=}0$) is assigned the fixed negative value $z_0$, and at evaluation a standardized prediction $\hat{z}$ is mapped back to physical units as $\hat{y}$.

\subsection{Backbone}
\label{sec:method-backbone}
The backbone is DINOv3 ViT-L/16~\cite{simeoni2025dinov3}, a plain Vision Transformer~\cite{dosovitskiy2021vit}
with embedding dimension $1024$, patch size $16$, $24$ blocks, rotary positional embeddings, and $4$ register
tokens~\cite{darcet2024registers}. Inputs are letterboxed to a square $896\times 896$ canvas, producing a
$56\times 56 = 3{,}136$ patch-token grid plus CLS and registers (\cref{fig:arch}). Letterbox padding
preserves aspect ratio, so the scalar labels remain unchanged under the D4 dihedral (eight flip and rotation) augmentations used in training and at inference (\cref{sec:method-train}). We keep the backbone
fully frozen, which preserves the transferable dense features that motivate the design (\cref{sec:related}); an
ablation (\cref{sec:exp-ablation}) indicates that unfreezing its top blocks does not improve performance.

\subsection{Hybrid parameter-efficient adaptation}
\label{sec:method-peft}
\begin{figure}[tbp]
\centering
\includegraphics[width=0.99\linewidth]{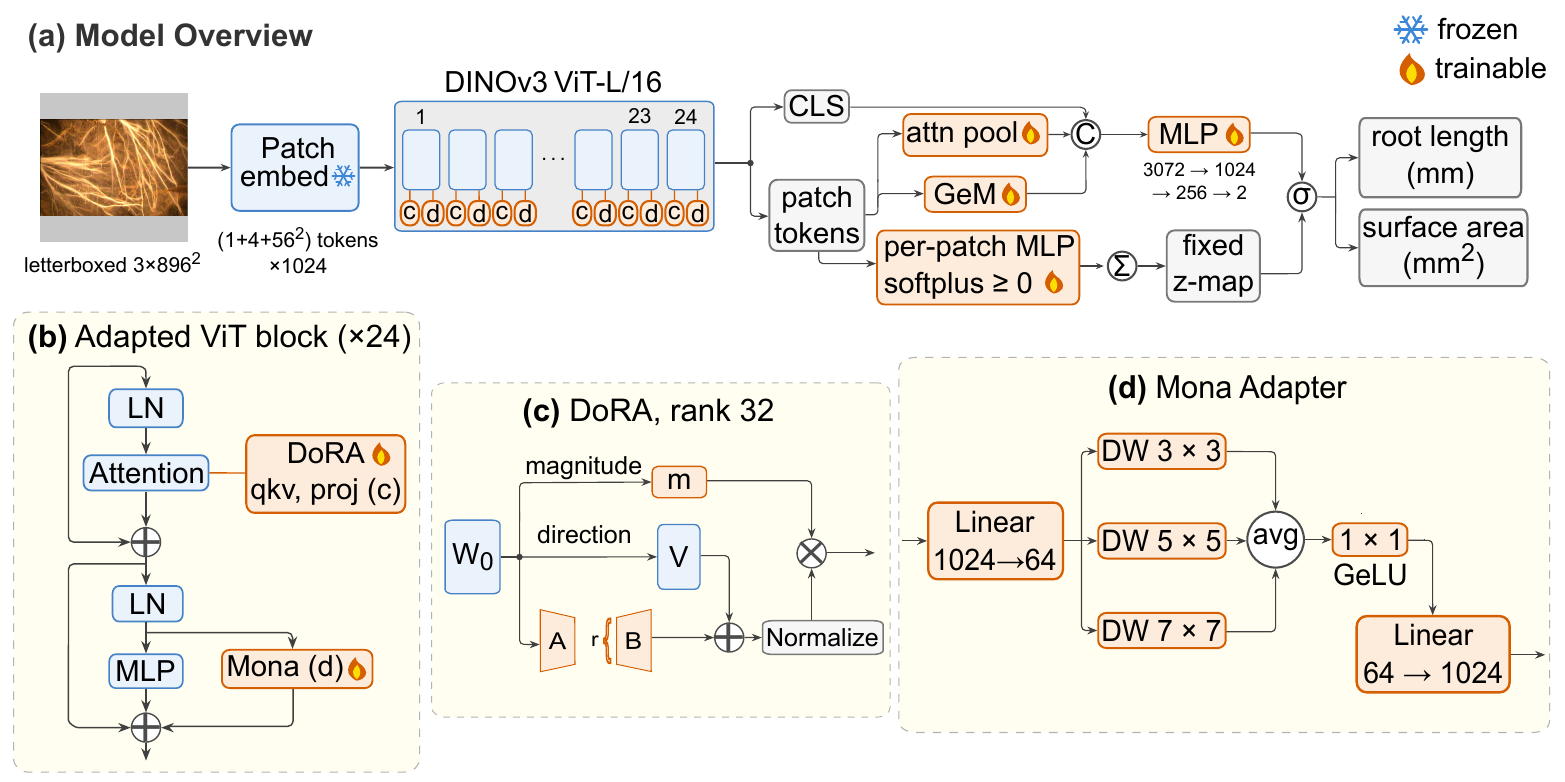}
\caption{\textbf{RootQuantV2 architecture.} Model with a frozen DINOv3 ViT-L/16 backbone, adapted ViT block, DoRA weight-decomposed low-rank adapter and Mona multi-scale convolutional adapter. Patch tokens are aggregated by CLS, attention, and GeM pooling, fused with an extensive density readout, and mapped to root length and surface area predictions.}
\label{fig:arch}
\end{figure}

We adapt the frozen backbone using two complementary mechanisms (\cref{fig:arch}).

\noindent\textbf{DoRA on attention.} Each block's fused query--key--value and output projections receive a
weight-decomposed low-rank update~\cite{liu2024dora}. DoRA decomposes a frozen weight $W$ into a per-output
magnitude and a direction, learns a magnitude vector $\mathbf{m}$ and a low-rank direction update $BA$
(rank $r=32$, $B$ zero-initialized), and recomposes
$W' = \mathbf{m} \, \frac{V + BA}{\lVert V + BA \rVert_{\mathrm{col}}}$, with $V = W/\lVert W\rVert_{\mathrm{col}}$
held fixed and $\lVert\cdot\rVert_{\mathrm{col}}$ the per-output-channel Euclidean norm. At $B{=}0$, $W'{=}W$, so
adaptation starts as an identity mapping.

\noindent\textbf{Mona on the MLP branch.} A flat transformer carries no two-dimensional locality, yet thin,
extended root structures benefit from one. In parallel to each block's MLP, we therefore add a multi-scale convolutional
adapter that injects this spatial inductive bias~\cite{yin2024mona}. We branch the adapter in parallel rather than
in series so the frozen MLP output reaches the residual undisturbed and the adapter only learns an additive
correction~\cite{chen2022adaptformer}. Patch tokens are down-projected from
$1024\to 64$, reshaped to a $56\times 56$ grid, and processed through depthwise convolutions of kernel sizes
$3\times3$, $5\times5$, and $7\times7$ whose outputs are averaged. The result is mixed by a $1\times1$ convolution and GELU, up-projected
from $64\to 1024$ using a zero-initialized weight, and regularized with dropout. The zero-initialized up-projection
makes the adapter a no-op at initialization, as in DoRA; a learnable scalar gate and per-sample stochastic depth
(rate $0.05$) then modulate the residual. CLS and register
tokens carry no spatial position and bypass the convolution. The hybrid pairs a channel-space update (DoRA) with a spatial update (Mona).

\noindent\textbf{Parameter budget.} The trainable budget is $11{,}904{,}545$
parameters, $3.78\%$ of the $315{,}058{,}737$-parameter model, split across DoRA ($4{,}816{,}896$), the Mona adapter ($3{,}403{,}800$), the regression head ($3{,}417{,}858$), the extensive readout ($264{,}966$), and the pooler ($1{,}025$; GeM exponent and attention query) only, with the DINOv3 backbone fully frozen.

\subsection{Regression-aware readout}
\label{sec:method-head}
We pool the patch tokens $t_i$ on the valid (non-padded) grid $\mathcal{V}$ using three complementary operators and concatenate them: (i) the CLS token, (ii) an attention pool that softmax-weights valid
tokens by their inner product with a single learnable query, and (iii) a generalized-mean pool (GeM) with a learnable
exponent applied to a non-negative (softplus-mapped) token map~\cite{radenovic2019gem}, since GeM presumes
non-negative activations whereas post-LayerNorm tokens are signed. The attention and GeM pools mask padded patch tokens, so only the CLS component---produced by the backbone over all tokens---retains any influence from the letterbox padding. The concatenated $3072$-d vector feeds a regression head that applies LayerNorm before each of two hidden layers ($3072\to1024$ and $1024\to256$, GELU and dropout), then a final $256\to2$ linear that outputs the standardized global prediction $\hat{z}^{\mathrm{glob}}$. In parallel, an extensive readout predicts a non-negative per-patch density, sums it over $\mathcal{V}$, and maps the total to standardized space with the \emph{fixed} present-root statistics of \cref{eq:standardize},
\begin{equation}
d_i = \operatorname{softplus}\!\big(\operatorname{MLP}_{\mathrm{dens}}(t_i)\big)\in\mathbb{R}^2_{\ge 0},
\qquad
\hat{z}^{\mathrm{dens}} = \frac{\mathbf{g}\odot \sum_{i\in\mathcal{V}} d_i - \boldsymbol{\mu}}{\boldsymbol{\sigma}},
\label{eq:density}
\end{equation}
where $\mathbf{g}$ is a learnable per-target gain and $\boldsymbol{\mu},\boldsymbol{\sigma}$ the fixed constants of \cref{eq:standardize}, so an empty frame ($\sum_{i\in\mathcal{V}} d_i\!\to\!0$) maps to $z_0$ exactly and the readout is calibrated to the empty target by construction. A learnable per-target gate then blends the extensive and the global (intensive) prediction,
\begin{equation}
\hat{z} = \alpha \odot \hat{z}^{\mathrm{glob}} + (1-\alpha)\odot \hat{z}^{\mathrm{dens}},
\qquad \alpha = \operatorname{sigmoid}(\beta)\in(0,1)^2,
\label{eq:blend}
\end{equation}
with logit $\beta$ initialized at $0$, so each trait starts from an equal blend of the two readouts and learns how far to lean on the extensive branch. The density branch receives no local supervision and is trained end-to-end using only the two global scalar targets. Summation therefore acts as an architectural inductive bias that matches the way root length and area accumulate with image content, rather than averaging over it. This design is inspired by count-regression and density-estimation approaches~\cite{lempitsky2010counting,zhang2016mcnn,liang2022transcrowd} only as motivation. We do not employ a density-map loss and make no claim of formal equivalence to object count.

\subsection{Targets, loss, and training}
\label{sec:method-train}
\begin{figure}[tbp]
\centering
\includegraphics[width=0.9\linewidth]{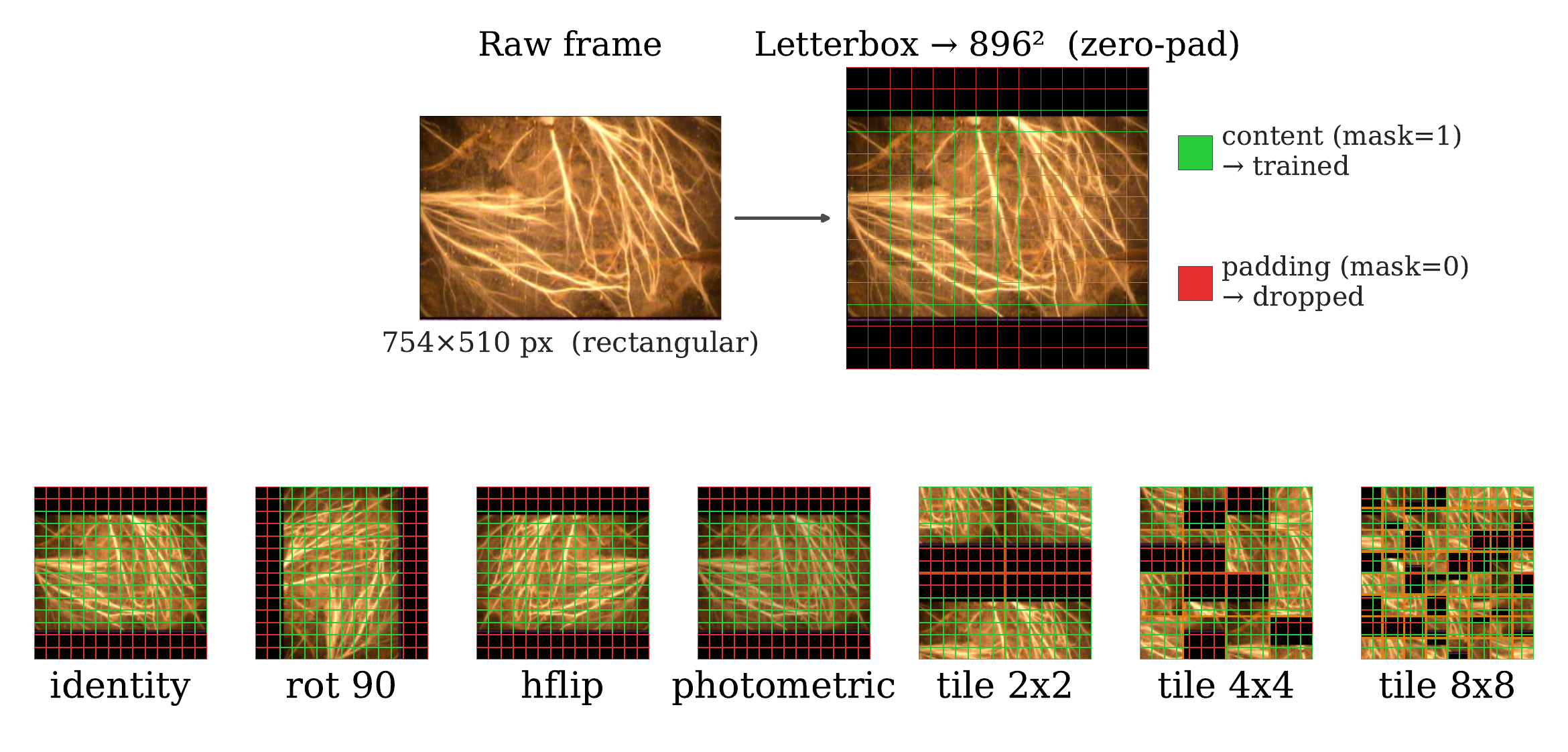}
\caption{\textbf{Input letterboxing and training augmentation.} Each frame (top-left) is letterboxed to an
$896\times 896$ square by resizing the long side to $896$ and zero-padding the short side (top-right). A per-patch
validity mask over the overlaid ViT-L/16 patch lattice ($56\times 56$ tokens) marks content tokens
(green, $\text{mask}=1$) and padding tokens (red, $\text{mask}=0$). Training (bottom) applies the D4 dihedral group
(eight flip and rotation views, green) and mild photometric jitter (gray); tile shuffle
(\cref{sec:method-train}) is a third augmentation.}
\label{fig:preprocess}
\end{figure}

We train the two standardized targets of \cref{eq:standardize} with a presence-balanced, task-weighted Huber loss,
\begin{equation}
\mathcal{L}=\frac{\sum_j w_j\,\mathcal{L}_j}{\sum_j w_j},
\qquad
\mathcal{L}_j=\frac{\sum_{y\in\{\ell,a\}}\lambda_y\,H_\delta\!\big(\hat{z}_{j,y}-z_{j,y}\big)}{\lambda_\ell+\lambda_a},
\label{eq:loss}
\end{equation}
where $j$ indexes images, $m_j$ is its presence indicator, $H_\delta$ is the Huber penalty~\cite{huber1964} ($\delta{=}3$, standardized units), $\lambda=(1,1.5)$ emphasizes the harder area target, and $w_j=\tfrac12[m_j/\rho+(1-m_j)/(1-\rho)]$ equalizes the total weight of present ($m_j{=}1$) and empty rows at present rate $\rho$. Optimization uses AdamW~\cite{loshchilov2019,kingma2015} with three parameter groups: DoRA and Mona adapters at $10^{-4}$, and the regression head and pooling layers at $10^{-3}$. The weight decay is $10^{-2}$, gradients are clipped to a norm of $1.0$, and the learning rate schedule follows cosine decay with $5\%$
warmup over $30$ epochs. All training is performed in full \texttt{fp32} precision. We train the full $30$ epochs without early stopping and use the validation split only to select the reported checkpoint, the epoch with the highest validation combined $R^2$. No configuration choice uses the test split.
Because $\ell$ and $a$ scale with the amount of root content, we restrict training augmentations to transformations that preserve these quantities. The D4 dihedral group---the eight flip and rotation symmetries---preserves both $\ell$ and $a$ exactly. Mild
photometric jitter (brightness, contrast, saturation, hue, and blur) also leaves the labels unchanged (\cref{fig:preprocess}). We further apply \emph{tile shuffle}, a
regularizer that partitions the letterboxed image into a $k\times k$ grid and permutes the tiles, disrupting global
layout while leaving intact the local root texture the Mona adapter models. The operation is applied independently
for $k\in\{2,4,8\}$ with probabilities $0.3$, $0.2$, and $0.1$, respectively. Since the permutation conserves
every root pixel, it approximately preserves the extensive targets while breaking global root layout and
discouraging the model from memorizing absolute position. We use no scale or crop augmentation, which would modify the targets. Finally, we maintain an exponential moving average of the trainable weights (decay
$0.9995$) and use the averaged weights for evaluation. Weight averaging improves generalization without increasing inference
cost~\cite{tarvainen2017meanteacher,izmailov2018swa,athiwaratkun2019fastswa}.

\subsection{Label-consistent inference}
\label{sec:method-infer}
At test time, we evaluate the exponential moving-average weights with a D4 augmentation ensemble. Each image is passed in
its eight dihedral views and the resulting $K=8$ predictions are averaged~\cite{shanmugam2021tta,wang2019tta}. Because D4
transformation leaves length and area invariant, the ensemble is label-consistent. No
back-transformation of the prediction is required, the average is taken in standardized space and inverted once by
\cref{eq:standardize}, and averaging incurs no bias-for-variance trade-off.
The idealized variance reduction of $1/K$ is achieved only when the per-view errors are uncorrelated~\cite{kimura2021tta}. All eight views share the same frozen backbone, so their errors are correlated and the realized reduction is smaller than $1/K$; we treat it as an empirical quantity measured against a matched no-augmentation baseline (\cref{sec:exp-ablation}).

\vspace{-1em}
\section{Experiments}
\label{sec:experiments}

\subsection{Setup and metrics}
\label{sec:exp-setup}
We train on $4\times$ A100 GPUs with a per-GPU batch of $8$ (effective batch $32$). For each target, we report the coefficient of
determination $R^2$, root-mean-square error (RMSE), and mean absolute error (MAE), computed over the $n$ images of a
subset in native units ($\ell$ in mm and $a$ in mm\textsuperscript{2}) after inverting \cref{eq:standardize}.
Combined $R^2_{\mathrm{comb}}=\tfrac12(R^2_\ell+R^2_a)$ is the mean of the per-target values, used only to rank configurations. Because RMSE is expressed in physical units, we do not average it across targets. RootQuant~\cite{rootquantv1} is evaluated on the same images using its per-image predictions, so both
models are compared on an identical test set. The strong zero-inflation makes full-set $R^2$ optimistic because the $12{,}692$ empty
images are relatively easy to fit. We therefore report
metrics over both the full test set ($17{,}313$ images, including empties) and the visible-root subset
(the $4{,}618$ images with $m{=}1$), which removes the inflation effect and is our primary indicator of trait recovery.

\vspace{-1em}
\subsection{Main Results}
\label{sec:exp-main}
\begin{figure}[!htbp]
\centering
\includegraphics[width=0.9\linewidth]{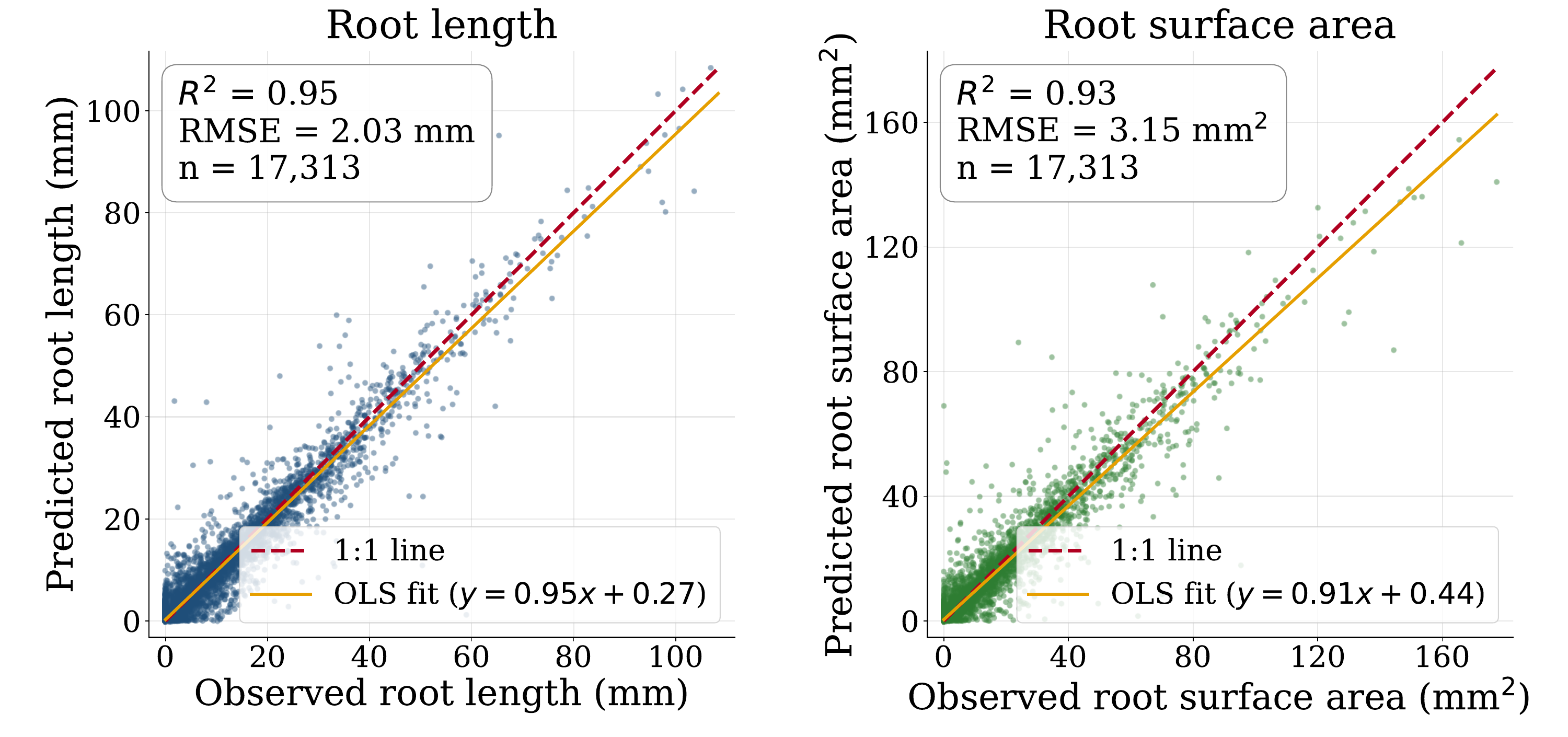}
\caption{Predicted-versus-true scatter plots for both traits using RootQuantV2 on the full test set ($n=17{,}313$, test-time-augmented). The dashed line denotes the 1:1 identity relationship.}
\label{fig:regression}
\end{figure}

On the visible-root subset, RootQuantV2 outperforms RootQuant across both length ($R^2$ increased by $6.6\%$) and area ($R^2$ increased by $6.4\%$), so the gain persists after removing the easy-to-predict empty images (\cref{tab:positive}). On the full
test set, the model achieves a combined $R^2$ of $0.940$, corresponding to a $+4.4\%$ improvement over RootQuant (\cref{tab:main}). Length and area RMSE decrease by $24.3\%$ and $20.7\%$, while MAE decreases by $23.8\%$ and $19.2\%$, respectively (\cref{tab:main}). 
The predicted-versus-true scatter plots (\cref{fig:regression}) show residuals widening with increasing trait magnitude, with a broader spread for area, consistent with its lower $R^2$. A single
mixed-species generalist serves both crops, reaching a full-set combined $R^2$ of $0.942$ for maize and $0.938$ for soybean; the per-species comparison against RootQuant is given on the visible-root subset (\cref{tab:positive}). All RootQuantV2 metrics use the label-consistent D4 ensemble, whose isolated contribution is quantified in \cref{sec:exp-ablation}.

\begin{table}[tbp]
\caption{Performance of RootQuant and RootQuantV2 on full test dataset ($n=17{,}313$), reported using $R^2$, RMSE, MAE, and the combined $R^2$ for root length (mm) and root surface area (mm$^2$). The lower rows report species-specific performance of RootQuantV2 on the maize and soybean subsets of the same test dataset.}
\label{tab:main}
\centering
\renewcommand{\arraystretch}{0.9}
\begin{tabular}{@{}lcccccc c@{}}
\toprule
& \multicolumn{3}{c}{Length} & \multicolumn{3}{c}{Area} & Comb.\\
\cmidrule(lr){2-4}\cmidrule(lr){5-7}
Model & $R^2$ & RMSE & MAE & $R^2$ & RMSE & MAE & $R^2$\\
\midrule
RootQuant (CNN) & 0.911 & 2.68 & 1.01 & 0.889 & 3.97 & 1.30 & 0.900\\
RootQuantV2 (ours) & \textbf{0.950} & \textbf{2.03} & \textbf{0.77} & \textbf{0.930} & \textbf{3.15} & \textbf{1.05} & \textbf{0.940}\\
\midrule
\quad maize subset & 0.959 & 1.42 & 0.60 & 0.925 & 2.59 & 0.84 & 0.942\\
\quad soybean subset & 0.945 & 2.39 & 0.90 & 0.931 & 3.52 & 1.20 & 0.938\\
\bottomrule
\end{tabular}
\end{table}

\begin{table}[tbp]
\caption{Performance of RootQuant and RootQuantV2 on the visible-root subset ($n=4{,}618$), reported using $R^2$, RMSE, MAE, and the combined $R^2$ for root length (mm) and root surface area (mm$^2$). The lower sections report species-specific performance on the maize and soybean subsets.}
\label{tab:positive}
\centering
\renewcommand{\arraystretch}{0.9}
\begin{tabular}{@{}lcccccc c@{}}
\toprule
& \multicolumn{3}{c}{Length} & \multicolumn{3}{c}{Area} & Comb.\\
\cmidrule(lr){2-4}\cmidrule(lr){5-7}
Model & $R^2$ & RMSE & MAE & $R^2$ & RMSE & MAE & $R^2$\\
\midrule
\multicolumn{8}{@{}l}{\emph{All visible ($n=4{,}618$)}}\\
RootQuant (CNN)~\cite{rootquantv1} & 0.866 & 5.04 & 3.13 & 0.850 & 7.51 & 4.14 & 0.858\\
RootQuantV2 (ours) & \textbf{0.923} & \textbf{3.81} & \textbf{2.26} & \textbf{0.904} & \textbf{6.01} & \textbf{3.12} & \textbf{0.914}\\
\midrule
\multicolumn{8}{@{}l}{\emph{Maize ($n=1{,}836$})}\\
RootQuant (CNN) & 0.930 & 3.13 & 2.04 & 0.879 & 5.84 & 3.03 & 0.904\\
RootQuantV2 (ours) & \textbf{0.947} & \textbf{2.72} & \textbf{1.70} & \textbf{0.905} & \textbf{5.16} & \textbf{2.53} & \textbf{0.926}\\
\midrule
\multicolumn{8}{@{}l}{\emph{Soybean ($n=2{,}782$)}}\\
RootQuant (CNN) & 0.829 & 5.98 & 3.85 & 0.831 & 8.43 & 4.87 & 0.830\\
RootQuantV2 (ours) & \textbf{0.908} & \textbf{4.39} & \textbf{2.63} & \textbf{0.899} & \textbf{6.51} & \textbf{3.51} & \textbf{0.904}\\
\bottomrule
\end{tabular}
\end{table}

\subsection{Development progression and ablations}
\label{sec:exp-ablation}
We trace test-set combined $R^2$ across three checkpoints, each adding capacity, locality, or resolution: (i) a $640$\,px DoRA baseline; (ii) a $768$\,px hybrid adding the Mona branch and the full recipe (GeM-softplus pool, weight averaging, Huber loss, extensive readout); (iii) the full $896$\,px model (DoRA $r{=}32$ plus Mona, frozen backbone), all evaluated with the eight-view D4 ensemble on the full test set (\cref{tab:progression}). The components of (ii) are introduced together and are not isolated individually. The full model reduces the baseline's full-set length MAE by $70.5\%$ ($2.61$ to $0.77$). Visible-root combined $R^2$ moves far less over the same progression ($0.905$ to $0.914$), so the full-set gain is dominated by empty-frame behavior. Unfreezing the MLP and normalization weights of the final two blocks adds $16.8$M trainable parameters ($28.7$M total, $9.11\%$ of the model), but degrades the full-set combined $R^2$ from $0.940$ to $0.936$. The frozen-backbone $11.9$M model is therefore the more parameter-efficient operating point and is used throughout the paper.

On identical full-model weights, the eight-view D4 average increases visible-root combined $R^2$ from $0.911$ to $0.914$ ($+0.3\%$; \cref{tab:mona}, $+$Mona row).

\begin{table}[tbp]
\caption{Performance evaluation throughout the development progression of RootQuantV2 on the full test set ($n=17{,}313$), with the
eight-view D4 ensemble, reported using $R^2$, RMSE, MAE, and the combined $R^2$ for root length (mm) and root surface area (mm$^2$).}
\label{tab:progression}
\centering
\renewcommand{\arraystretch}{0.9}
\begin{tabular}{@{}lcccccc c@{}}
\toprule
& \multicolumn{3}{c}{Length} & \multicolumn{3}{c}{Area} & Comb.\\
\cmidrule(lr){2-4}\cmidrule(lr){5-7}
Configuration & $R^2$ & RMSE & MAE & $R^2$ & RMSE & MAE & $R^2$\\
\midrule
DoRA baseline ($640$\,px, $r{=}16$) & 0.874 & 3.20 & 2.61 & 0.889 & 3.96 & 2.72 & 0.882\\
\;+ Mona, recipe ($768$\,px) & 0.945 & 2.12 & 0.83 & 0.923 & 3.30 & 1.05 & 0.934\\
Full \;($896$\,px, $r{=}32$, frozen; $11.9$M) & \textbf{0.950} & \textbf{2.03} & \textbf{0.77} & \textbf{0.930} & \textbf{3.15} & \textbf{1.05} & \textbf{0.940}\\
\;\;\;+ unfreeze last $2$ ($28.7$M) & 0.947 & 2.08 & 0.79 & 0.924 & 3.27 & 1.02 & 0.936\\
\bottomrule
\end{tabular}
\end{table}

\noindent\textbf{The matched ablation shows that Mona contributes primarily on the area trait.} Both arms use a frozen DINOv3 backbone at $896$\,px with DoRA ($r{=}32$) applied to attention layers; the only difference is the Mona MLP-branch adapter ($+3.40$M parameters). Mona is therefore the sole adapter on the MLP branch. The Mona-on arm is the model we report throughout. Adding Mona reduces visible-root area RMSE by $9.0\%$ without augmentation ($R^2$ $0.883\to 0.903$) and by $5.5\%$ with the D4 ensemble ($R^2$ $0.893\to 0.904$) (\cref{tab:mona}). Combined $R^2$ increases by $+1.2\%$ ($0.900\to 0.911$) without augmentation and by $0.8\%$ ($0.907\to 0.914$) with augmentation, while length metrics improve only marginally. The largest single gain is soybean area, the previously laggard trait ($R^2$ $0.878\to0.901$ without augmentation, $0.888\to0.899$ with the D4 ensemble; \cref{tab:mona}).

\begin{table}[tbp]
\caption{Performance evaluation of the RootQuantV2 DoRA/Mona ablation pair on the visible-root subset, by species and inference mode (no-augmentation vs.\ the eight-view D4 ensemble, TTA). $n=4{,}618$ (maize $1{,}836$, soybean $2{,}782$). The $+$Mona configuration is the main RootQuantV2 model.}
\label{tab:mona}
\centering
\renewcommand{\arraystretch}{0.9}
\setlength{\tabcolsep}{4pt}
\renewcommand{\arraystretch}{0.9}
\footnotesize
\begin{tabular}{@{}llcccccc c@{}}
\toprule
& & \multicolumn{3}{c}{Length} & \multicolumn{3}{c}{Area} & Comb.\\
\cmidrule(lr){3-5}\cmidrule(lr){6-8}
Subset & Configuration & $R^2$ & RMSE & MAE & $R^2$ & RMSE & MAE & $R^2$\\
\midrule
\multirow{4}{*}{All vis.}
 & DoRA only, no-aug & 0.916 & 3.99 & 2.39 & 0.883 & 6.63 & 3.27 & 0.900\\
 & DoRA only, TTA    & 0.921 & 3.87 & 2.33 & 0.893 & 6.36 & 3.18 & 0.907\\
 & $+$Mona, no-aug  & 0.919 & 3.93 & 2.33 & 0.903 & 6.03 & 3.16 & 0.911\\
 & $+$Mona, TTA     & \textbf{0.923} & \textbf{3.81} & \textbf{2.26} & \textbf{0.904} & \textbf{6.01} & \textbf{3.12} & \textbf{0.914}\\
\midrule
\multirow{4}{*}{Maize}
 & DoRA only, no-aug & 0.939 & 2.92 & 1.85 & 0.884 & 5.71 & 2.72 & 0.912\\
 & DoRA only, TTA    & 0.941 & 2.86 & 1.80 & 0.893 & 5.48 & 2.64 & 0.917\\
 & $+$Mona, no-aug  & 0.946 & 2.75 & 1.75 & 0.900 & 5.31 & 2.58 & 0.923\\
 & $+$Mona, TTA     & \textbf{0.947} & \textbf{2.72} & \textbf{1.70} & \textbf{0.905} & \textbf{5.16} & \textbf{2.53} & \textbf{0.926}\\
\midrule
\multirow{4}{*}{Soybean}
 & DoRA only, no-aug & 0.901 & 4.55 & 2.75 & 0.878 & 7.17 & 3.62 & 0.890\\
 & DoRA only, TTA    & 0.907 & 4.42 & 2.68 & 0.888 & 6.88 & 3.54 & 0.897\\
 & $+$Mona, no-aug  & 0.901 & 4.54 & 2.70 & \textbf{0.901} & \textbf{6.46} & 3.55 & 0.901\\
 & $+$Mona, TTA     & \textbf{0.908} & \textbf{4.39} & \textbf{2.63} & 0.899 & 6.51 & \textbf{3.51} & \textbf{0.904}\\
\bottomrule
\end{tabular}
\end{table}

\subsection{Cross-species transferability}
\label{sec:exp-xspecies}
The headline model is a single \emph{mixed-species generalist} trained jointly on maize and soybean. Starting from a model trained exclusively on soybean, we evaluated three transfer regimes (\cref{tab:xspecies}, visible-root subset, test-time-augmented): (i) zero-shot transfer (ZST: soybean$\to$soybean as an in-domain reference, and soybean$\to$maize with no adaptation); (ii) frozen cross-fine-tune (FCFT: soybean$\to$maize with only the readout head retrained on maize), and (iii) full fine-tune (FFT: soybean$\to$maize with all trainable parameters adapted). The soybean-only model is a strong in-domain regressor, achieving combined $R^2$ $0.903$ on soybean, but zero-shot transfer on maize drops combined $R^2$ by $12.2\%$ to $0.793$. Retraining only the head (FCFT) recovers roughly one-third of the gap to full fine-tuning (combined $R^2$ $0.843$) at negligible additional cost. When the \emph{same} fully fine-tuned (FFT) model is re-evaluated on soybean, combined $R^2$ drops by $17.9\%$ to $0.741$, indicating substantial forgetting of the source domain.

The mixed-species generalist reaches $R^2$ $0.947/0.905$ length/area on the maize visible-root subset (\cref{tab:positive}), close to the maize-specialized FFT model's $R^2$ $0.954/0.910$, and matches the soybean-only model on soybean (combined $R^2$ $0.904$ vs.\ $0.903$; \cref{tab:positive,tab:xspecies}). It is therefore on par with the in-domain specialist while also covering maize, which is why we adopt it rather than per-species specialists.

\begin{table}[tbp]
\caption{Cross-species transfer of a \emph{soybean-only} model vs.\ the mixed-species generalist, on the visible-root subset ($n=4{,}618$) with D4 TTA, reported using $R^2$, RMSE, and MAE for root length (mm) and root surface area (mm$^2$). All regimes warm-start from the soybean-only model; \emph{Eval} is the test species. ZST: zero-shot (no maize adaptation); FCFT: readout head fine-tuned on maize; FFT: all weights fine-tuned on maize. \emph{(forgetting)} rows re-evaluate each maize-tuned model on soybean.}
\label{tab:xspecies}
\centering
\renewcommand{\arraystretch}{0.9}
\setlength{\tabcolsep}{3pt}
\renewcommand{\arraystretch}{0.9}
\footnotesize
\begin{tabular}{@{}llcccccc c@{}}
\toprule
& & \multicolumn{3}{c}{Length} & \multicolumn{3}{c}{Area} & Comb.\\
\cmidrule(lr){3-5}\cmidrule(lr){6-8}
Regime & Eval & $R^2$ & RMSE & MAE & $R^2$ & RMSE & MAE & $R^2$\\
\midrule
ZST (in-domain)     & soybean & 0.911 & 4.31 & 2.63 & 0.896 & 6.63 & 3.52 & 0.903\\
ZST (cross-species) & maize   & 0.814 & 5.10 & 2.86 & 0.772 & 8.01 & 3.98 & 0.793\\
FCFT (head on maize) & maize  & 0.881 & 4.08 & 2.70 & 0.805 & 7.42 & 4.39 & 0.843\\
\quad\emph{(forgetting)} & soybean & 0.658 & 8.46 & 5.70 & 0.567 & 13.51 & 9.12 & 0.612\\
FFT (full on maize) & maize   & \textbf{0.954} & \textbf{2.54} & \textbf{1.54} & \textbf{0.910} & \textbf{5.02} & \textbf{2.38} & \textbf{0.932}\\
\quad\emph{(forgetting)} & soybean & 0.712 & 7.77 & 4.26 & 0.770 & 9.84 & 5.52 & 0.741\\
\midrule
\multirow{2}{*}{Generalist (mixed)} & soybean & 0.908 & 4.39 & 2.63 & 0.899 & 6.51 & 3.51 & 0.904\\
                                    & maize   & 0.947 & 2.72 & 1.70 & 0.905 & 5.16 & 2.53 & 0.926\\
\bottomrule
\end{tabular}
\end{table}

\vspace{-1em}

\subsection{Feature saliency and trait density}
\label{sec:exp-interp}
\begin{figure}[tbp]
\centering
\includegraphics[width=0.85\linewidth,clip,trim=0 0 0 1]{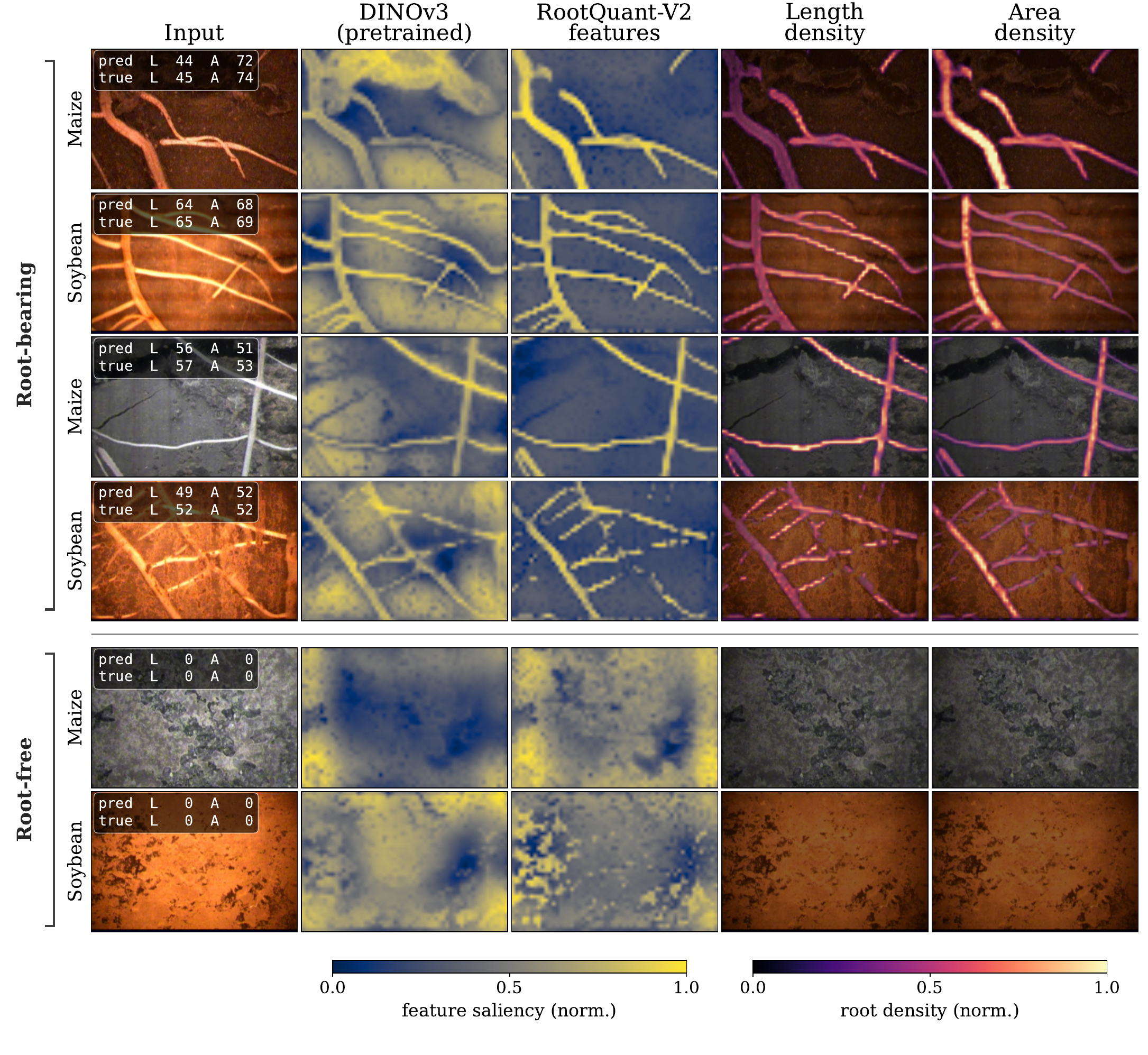}

\caption{\textbf{Feature saliency and trait density maps on representative test images.} The upper panel shows root-bearing frames, and the lower panel shows root-free frames. Input images (left), task-agnostic feature saliency computed from the variance-weighted magnitude of the top three principal components of patch tokens for off-the-shelf DINOv3 and for RootQuantV2 (center), and task-faithful per-patch density maps for root length and root surface area produced by the extensive readout (right) (\cref{sec:exp-interp}). Predicted and true values for $(\text{Length (L)}, \text{Area (A)})$ are overlaid at the top-left of each input.}
\label{fig:saliency}
\end{figure}

Localization of root structure is largely inherited from pretraining. We show this for the headline model with two types of per-patch map, both obtained from a single forward pass on the letterbox input without the D4 ensemble (\cref{fig:saliency}). \textbf{Feature saliency.} We define feature saliency as the variance-weighted magnitude of the top three principal components (PCs) of the token representations. Because no task labels are used in PCA, this map is \emph{task-agnostic}. The same PCA basis is applied to two representations: (i) off-the-shelf \emph{DINOv3 (pretrained)} tokens, and (ii) \emph{RootQuantV2 tokens} produced by the same backbone with DoRA and Mona active in the forward pass. \textbf{Trait density.} The length and area density maps are the per-patch densities $d_i$ of the extensive readout (\cref{eq:density}), shown before summation, with one map per trait. These maps are \emph{task-faithful}, since their masked sum over the valid patches equals the corresponding global prediction. Off-the-shelf DINOv3 already localizes the root structures. The DoRA and Mona adaptations primarily increase contrast---root responses become sharper while the background soil substrate is suppressed, with little change in the spatial location of the salient regions (\cref{fig:saliency}).

The density maps localize on roots rather than on the surrounding substrate despite receiving no local supervision. On root-bearing frames, they trace the root structures for both species, and their masked sum recovers the predicted trait value. On root-free frames whose substrate carries root-like texture, the density maps remain near zero, whereas the task-agnostic feature saliency still highlights parts of that texture. 
This behavior emerges from the image-level objective and explains the model's most frequent failure, a prediction above $1$\,mm (length) or $1$\,mm\textsuperscript{2} (area) on $4.5\%$ and $4.1\%$ of true root-free images, respectively, where root-like substrate texture is not fully suppressed. Across all $12{,}692$ root-free frames the predictions stay near zero (medians $0.10$\,mm and $0.18$\,mm\textsuperscript{2}, $95$th percentiles $0.90$\,mm and $0.85$\,mm\textsuperscript{2}), summing to $5.1\%$ and $5.6\%$ of the true visible-root length and area. These false positives are frequent but metric-cheap. An oracle presence gate zeroing them would increase combined $R^2$ by only $0.002$, because $R^2$ is variance-weighted and dominated by the visible-root fit. The large full-set gain over the $640$\,px baseline (\cref{sec:exp-ablation}) therefore comes from suppressing that baseline's much larger empty-frame errors, not from the small residual false positives that remain.

\vspace{-1em}
\section{Discussion}
\label{sec:discussion}

Parameter-efficient adaptation of a frozen self-supervised vision foundation model substantially improves segmentation-free root-trait regression. The adapted ViT outperforms the convolutional RootQuant backbone on minirhizotron imagery, although the two models also differ in input resolution, readout, loss, and inference, so the comparison is not backbone-controlled. Off-the-shelf DINOv3 already localizes root structure without supervision (\cref{fig:saliency}), yet its features cannot regress length and area without tuning and a trait-matched readout. DoRA adapts channel space while preserving the pretrained directional structure, and Mona adds multi-scale convolution to the MLP branch, restoring spatial inductive bias to the flat transformer. The matched ablation shows Mona acts mainly on the more spatially distributed area trait, cutting visible-root area RMSE by $9.0\%$ without test-time augmentation and $5.5\%$ with it, while length improves only marginally. Area depends on both length and diameter, which a multi-scale convolution may resolve better, although we do not measure diameter and cannot test this directly. As length and area are extensive totals, not pixel-level labels, the extensive readout sums a learnable per-patch density across the grid (\cref{eq:density}), so each trait is the integral of allocated evidence, and a learned per-target gate (\cref{eq:blend}) sets how far each trait leans on that branch; concatenating it with CLS, attention-pool, and GeM-pool tokens may also contribute to the consistent accuracy on empty and visible-root frames. The $70.5\%$ drop in full-set length MAE over the baseline (\cref{sec:exp-ablation}) tracks a fall in the frequency and magnitude of false positives on empty frames, and the density maps are consistent with that account, placing evidence predominantly where roots are present rather than on root-like soil texture, though not on every frame (\cref{sec:exp-interp}).

The mixed-species generalist matches in-domain specialist performance, yet zero-shot transfer between species drops substantially (\cref{sec:exp-xspecies}), showing that features learned for one species do not fully transfer; head fine-tuning recovers part of the gap, while full adaptation risks catastrophic forgetting. The generalist was trained on both species from the start; whether a pretrained backbone can instead adapt to a new crop from only a few hundred labeled examples remains to be validated. Those labels are the same numeric totals the existing archives already contain, so such fine-tuning would need no new annotation.

\vspace{-1em}
\section{Conclusion}
\label{sec:conclusion}

RootQuantV2 recovers both root traits with $R^2$ above $0.9$, outperforming the convolutional RootQuant baseline by $4.4$--$6.5\%$ while training only $3.78\%$ of its parameters. It attains this by adapting a frozen self-supervised vision foundation model with a hybrid DoRA--Mona adaptation scheme and an extensive readout matched to root length and area. Because it needs only the numeric length and area archives already produced by decades of manual tracing, the method repurposes legacy data to scale root phenotyping to large image collections without further annotation. A single mixed-species generalist covers both maize and soybean, suggesting a path toward rapid deployment on new crops with limited labeled data.

\vspace{-1em}
{\small
\subsubsection*{Acknowledgements.}
We thank all members of the Leakey Laboratory for the field trials, the data collection, and above all the manual root tracing on this dataset from 2009 to 2020, without which this work would not have been possible. Funded by the National Science Foundation Plant Genome Research Program (award IOS-1638507); the Advanced Research Projects Agency--Energy (ARPA-E), U.S.\ Department of Energy (award DE-AR0000661); the DOE Center for Advanced Bioenergy and Bioproducts Innovation (Office of Science, Biological and Environmental Research Program, award DE-SC0018420); the Artificial Intelligence for Future Agricultural Resilience, Management, and Sustainability (AIFARMS) Institute (USDA National Institute of Food and Agriculture, Agriculture and Food Research Initiative grant no.\ 2020-67021-32799, project accession no.\ 1024178); and a generous gift from Tito's Handmade Vodka.
\par}

\bibliographystyle{splncs04}
\bibliography{main}
\end{document}